\documentclass[research]{fcs}

\usepackage{graphicx} 
\usepackage{tabularx}
\usepackage{multirow}
\usepackage{pifont}
\usepackage{threeparttable}

\usepackage{microtype}

\usepackage[capitalize]{cleveref}
\crefname{section}{Sec.}{Secs.}
\Crefname{section}{Section}{Sections}
\Crefname{table}{Table}{Tables}
\crefname{table}{Tab.}{Tabs.}

\def\algo{{\textsc{RoLT}}}

\def\eqref#1{equation~\ref{#1}}

\def\1{\bm{1}}

\def\algo{{\textsc{RoLT}}}

\def\c{{\boldsymbol c}}
\def\g{{\boldsymbol g}}
\def\x{{\boldsymbol x}}
\def\y{{\boldsymbol y}}

\def\DD{{\mathcal D}}

\def\LL{{\mathcal L}}
\def\SS{{\mathcal S}}
\def\XX{{\mathcal X}}

\newcommand{\cmark}{\ding{51}}%
\newcommand{\xmark}{\ding{55}}%

\definecolor{Gray}{gray}{0.9}
\definecolor{LightCyan}{rgb}{0.88,1,1}

\newcolumntype{a}{>{\columncolor{Gray}}c}
\newcolumntype{b}{>{\columncolor{white}}c}

\newcommand*{\belowrulesepcolor}[1]{%
  \noalign{%
    \kern-\belowrulesep 
    \begingroup 
      \color{#1}%
      \hrule height\belowrulesep 
    \endgroup 
  }%
} 
\newcommand*{\aboverulesepcolor}[1]{%
  \noalign{%
    \begingroup 
      \color{#1}%
      \hrule height\aboverulesep 
    \endgroup 
    \kern-\aboverulesep 
  }%
}

\title{Robust Long-Tailed Learning Under Label Noise}

\author{Tong Wei$^{1,2,3*}$}
\author{Jiang-Xin Shi$^{3,4*}$}
\author{Min-Ling Zhang$^{1,2}$}
\author*{Yu-Feng Li$^{3,4}$}
\address[1]{School of Computer Science and Engineering, Southeast University, Nanjing 210096, China}
\address[2]{Key Laboratory of Computer Network and Information Integration (Southeast University), Ministry of Education, China}
\address[3]{National Key Laboratory for Novel Software Technology, Nanjing University, Nanjing 210023, China}
\address[4]{School of Artificial Intelligence, Nanjing University, Nanjing 210023, China}
\fcssetup{
  received       = {month dd, yyyy},
  accepted       = {month dd, yyyy},
  corr-email     = {liyf@lamda.nju.edu.cn \\ $^*$ Tong Wei and Jiang-Xin Shi contributed equally to this work.},
}

\begin{abstract}
Long-tailed learning aims to enhance the generalization performance of underrepresented tail classes. However, previous methods have largely overlooked the prevalence of noisy labels in training data. In this paper, we address the challenge of noisy labels in long-tailed learning. We identify a critical issue: the commonly used small-loss noisy label detection criterion fails to perform effectively in long-tailed class distributions. This failure arises from the inherent bias of deep neural networks, which tend to misclassify tail class examples as head classes, leading to unreliable loss calculations. To mitigate this, we propose a novel small-distance criterion that leverages the robustness of learned representations, enabling more accurate identification of correctly-labeled examples across both head and tail classes. Additionally, to improve training for tail classes, we replace discrete pseudo-labels with label distributions for examples flagged as noisy, resulting in significant performance gains. Based on these contributions, we introduce the \textit{Robust Long-Tail learning} framework, designed to train models that are resilient to both class imbalance and noisy labels. Extensive experiments on benchmark and real-world datasets demonstrate that our approach outperforms previous methods, offering substantial performance improvements.
\end{abstract}

\begin{document}

\section{Introduction}


Classification problems in real-world typically exhibit a long-tailed class distribution, where most classes are associated with only a few examples, e.g., visual recognition~\cite{DBLP:conf/cvpr/HornASCSSAPB18,DBLP:conf/cvpr/0002MZWGY19,DBLP:conf/cvpr/TanWLLOYY20}, instance segmentation~\cite{DBLP:conf/cvpr/GuptaDG19}, and text categorization~\cite{DBLP:conf/nips/VaswaniSPUJGKP17,weit2020tnnls}. Due to the paucity of training examples, generalization for tail classes is challenging; moreover, na\"{\i}ve learning on such data is susceptible to an undesirable bias towards head classes. Recently, \textit{long-tail learning (LTL)} has gained renewed interest in the community~\cite{DBLP:conf/icml/CardieN97,zhou2005training,he2009learning,DBLP:conf/nips/WangRH17,DBLP:conf/cvpr/CuiJLSB19,wang2021diverse,wei2021towards,wei2023towards,shi2024long}. Two active strands of work involve normalisation of the classifier's weights, and modification of the underlying loss to account for different class penalties. Each of these strands is intuitive, and has been empirically shown to be effective~\cite{DBLP:conf/iclr/KangXRYGFK20,DBLP:conf/eccv/WuH0WL20,DBLP:journals/corr/abs-2104-02703}.

Existing LTL methods with remarkable performance are mostly trained on \textit{clean datasets} with high-quality human annotations. However, in real-world machine learning applications, annotating a large-scale dataset is costly and time-consuming. Some recent works resort to a large amount of web data as a source of supervision for training deep neural networks~\cite{webvision}. While the existing works have shown advantages in various applications~\cite{DBLP:conf/eccv/LiNX14,ma2018dimensionality}, web data is naturally under long-tailed class distribution accompanied with noisy labels~\cite{xu2019l_dmi,yao2020dualt,xia2021robust,li2021mopro}. As a result, it is crucial that deep neural networks can harvest noisy and long-tailed training data. Nevertheless, deep neural networks (DNNs) are prone to overfitting noisy labels. This problem has been widely studied in the literature~\cite{zhang2017understanding} on learning from noisy labels. Without considering noisy labels, we show that LTL methods severely degrade their performance in experiments.

\label{sec:introduction}

\begin{figure}[t]
    \centering
\includegraphics[width=0.9\linewidth]{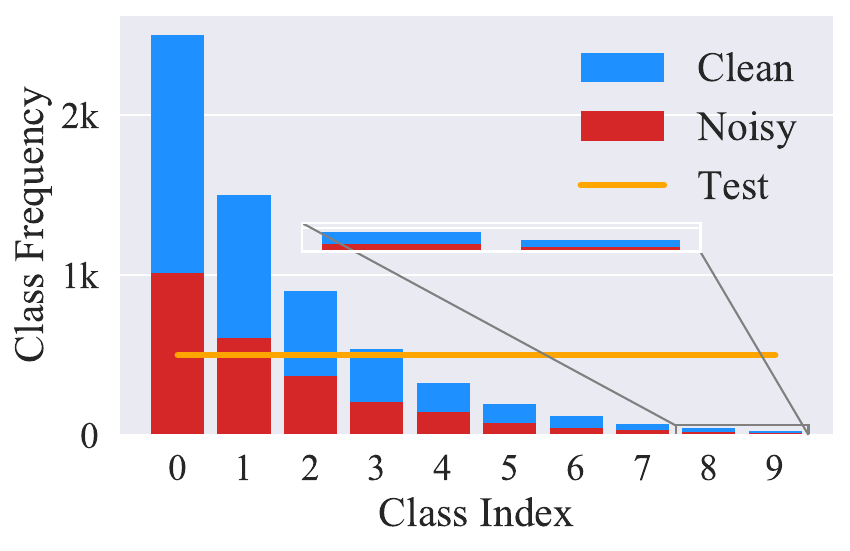}
\caption{Problem setting illustration.}\label{fig:problem_setup}
\end{figure}

\begin{figure*}[!t]
    \centering
    \begin{subfigure}[b]{0.23\textwidth}
        \centering
        \includegraphics[width=\textwidth]{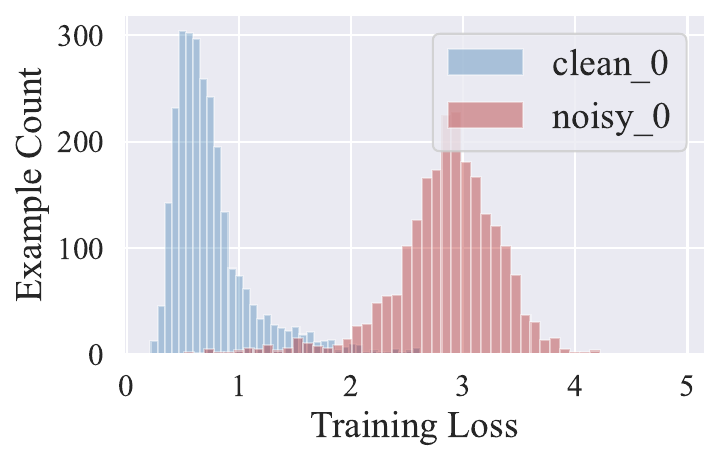}
        \caption{ loss of head class}
        \label{fig:loss_class_0}
    \end{subfigure}
    \begin{subfigure}[b]{0.23\textwidth}
        \centering
        \includegraphics[width=\textwidth]{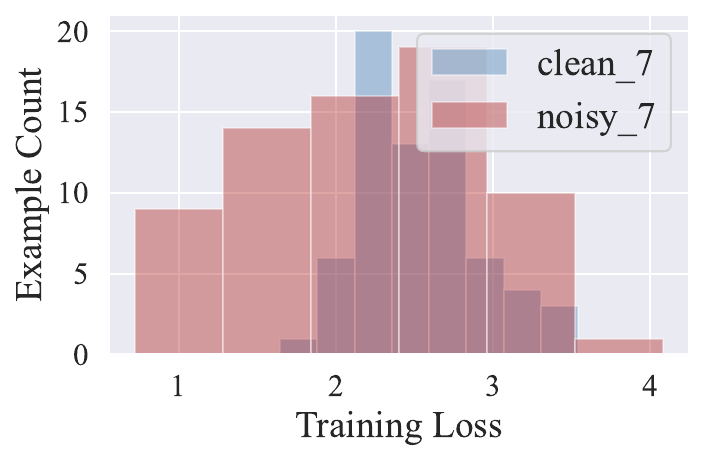}
        \caption{ loss of tail class}
        \label{fig:loss_class_7}
    \end{subfigure}
    \begin{subfigure}[b]{0.25\textwidth}
        \centering
        \includegraphics[width=\textwidth]{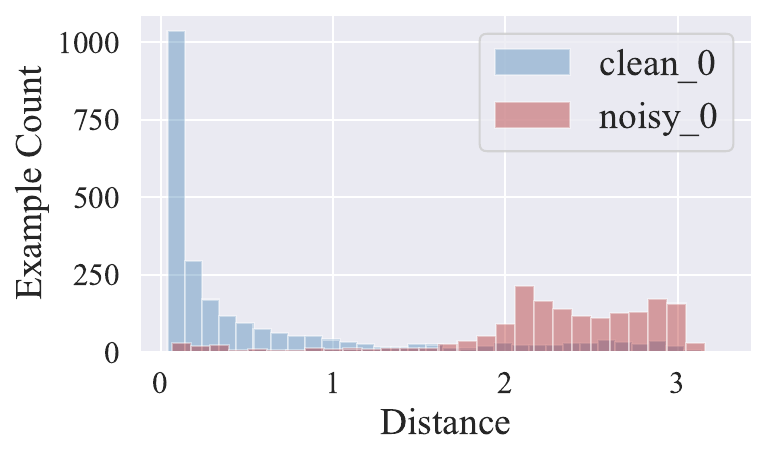} 
        \caption{distance of head class}\label{fig:distance_class_0}
    \end{subfigure}
    \begin{subfigure}[b]{0.25\textwidth}
        \centering
        \includegraphics[width=\textwidth]{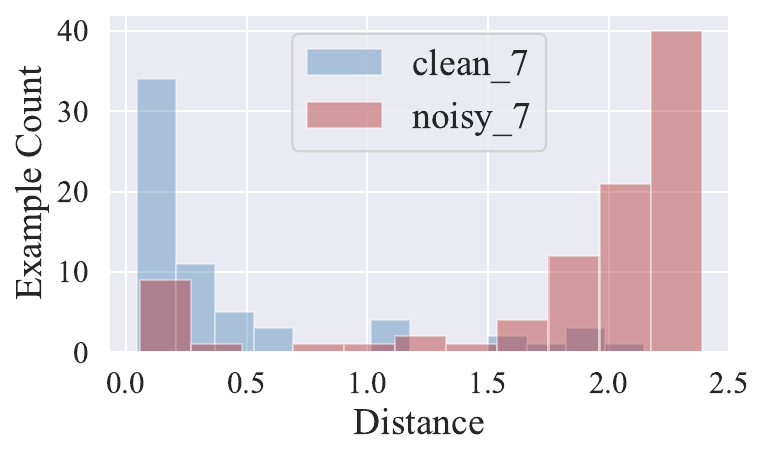} 
        \caption{distance of tail class}\label{fig:distance_class_7}
    \end{subfigure}
    \caption{(a-b) Training losses head and tail class samples. (c-d) Distance distribution between samples and their class prototype for head and tail class. Experiments are conducted on CIFAR-10 with noise level $\gamma=50\%$ and imbalance ratio $\rho=100$.}
\end{figure*}

In this paper, we investigate the problem of learning from noisy and long-tailed data, which is a more realistic setting but still underexplored. We provide a simple visualization of the studied problem in Figure~\ref{fig:problem_setup}. 
To reduce the negative impact of noisy labels, learning with noisy labels has gained a lot of attention in recent years and a lot of approaches have been proposed~\cite{natarajan2013learning,frenay2013classification,xu2019l_dmi,yao2020dualt,xia2021robust,wang2021learning,song2022survey}. Existing works can be roughly divided into two strands, i.e., noise transition matrix estimation~\cite{liu2016importance,hendrycks2018using} and sample selection~\cite{wang2018iterative,yao2020searching,lee2019rog}. Since the noise transition matrix is hard to be estimated especially when the number of classes is large, sample selection is a more promising way of handling noisy labels and is our focus in this paper. In sample selection methods, the \textit{small-loss} criterion is one of the most popular approaches~\cite{jiang2018mentornet,han2018co,arazo2019unsupervised,li2020dividemix,xia2022sample}. It selects samples with small losses and treats these samples as correctly annotated for robust training. In recent years, the small-loss criterion has been demonstrated to be effective in many works.

However, the small-loss criterion selects possibly clean samples using a constant threshold~\cite{han2018co,xia2022sample} or mixture distribution model~\cite{arazo2019unsupervised,li2020dividemix} for all classes, thus failing to consider different learning status and learning difficulties of different classes, which is very important in LTL. Owing to this paucity of samples for tail classes, na\"{\i}ve method is susceptible to an undesirable bias towards head classes. Specifically, due to the misclassification of samples with tail classes (large losses), the small-loss criterion cannot distinguish clean samples of tail classes from samples with noisy labels. Once the wrong selection is made, the inferiority of accumulated errors will arise. We further confirm this standpoint by experiments as shown in Figure~\ref{fig:loss_class_0} and~\ref{fig:loss_class_7}.


When handling noisy labels in long-tailed data, we believe it is important to keep the approach simple and free of auxiliary supervision. The benefit of doing this is that the approach can be easily absorbed by many existing frameworks for learning with long-tailed data. Guided by this belief, we propose the \textit{class-wise small-distance} criterion. For each individual class, it selects small-distance samples as clean where distance is calculated between the sample and its class prototype in the embedding space. The intuition why the \textit{class-wise small-distance} criterion can be more robust than \textit{small-loss} is briefly explained as follows:
(1) \textit{why is sample-distance better than small-loss?} As confirmed by many previous literature~\cite{lee2019rog,wu2020topofilter,Wu_2021_ICCV,DBLP:journals/pami/TangPHPC23}, it is reasonable to assume that clean examples tend to be clustered around their prototypes even when training with noisy labels.
(2) \textit{why can the sample selection work well in a class-wise manner?} As the number of classes can be large and the population of classes varies significantly in training data, the variance of distances between samples and class prototypes becomes large. We show the distance distribution for both head and tail classes in Figure~\ref{fig:distance_class_0} and~\ref{fig:distance_class_7}. Moreover, the proposed \textit{class-wise small-distance} is general and can be combined with semi-supervised learning to improve generalization.


\textbf{Our main contributions} are summarized as follows.
    (1) We address the underexplored problem of learning from long-tail data in the presence of noise, taking a significant step towards real-world applications.
(2) We identify the limitations of the widely used small-loss criterion under long-tailed class distributions and introduce a novel criterion called \textit{class-wise small-distance} to overcome this challenge. Building upon this, we propose a robust framework called \algo, which incorporates label distributions to enhance training for tail classes. Additionally, we present an improved version, \algo+, which can be seamlessly integrated with existing semi-supervised learning methods with minimal additional overhead. (3) Extensive experiments conducted on benchmark and real-world datasets demonstrate the superiority of our proposed method.

\section{Related Work}

\textbf{Long-tail learning.} Recently, many approaches have been proposed in LTL~\cite{DBLP:conf/nips/YangX20rethinking,zhang2020towards,DBLP:journals/corr/abs-2104-02703,wang2021diverse}. Most extant approaches can be categorized into three types by modifying (i) the inputs to a model by re-balancing the training data~\cite{DBLP:conf/eccv/ShenLH16,DBLP:conf/cvpr/0002MZWGY19,DBLP:conf/cvpr/ZhouCWC20bbn}; (ii) the outputs of a model, for example by post-hoc adjustment of the classifier~\cite{DBLP:journals/corr/abs-2001-01385,DBLP:conf/iclr/KangXRYGFK20,tang2020tde,menon2021logitadjust}; and (iii) the internals of a model by modifying the loss function~\cite{DBLP:conf/nips/CaoWGAM19ldam,DBLP:conf/nips/ShuXY0ZXM19,DBLP:conf/cvpr/CuiJLSB19,DBLP:conf/eccv/WuH0WL20,jamal2020rethinking,DBLP:conf/nips/RenYSMZYL20}. Each of the above methods are intuitive, and have shown strong empirical performance. However, these methods assume the training examples are correctly-labeled, which is often difficult to obtain in many real-world applications. Instead, we study a realistic problem to learn from long-tailed data with label noise. Although the presence of label noise in class-imbalanced datasets has also been mentioned in HAR~\cite{cao2021heteroskedastic}, they only consider a specific setup. In this work, we provide systematic studies concerning noisy labels in LTL.

\noindent
\textbf{Label noise detection.} Plenty of methods have been proposed to detect noisy examples~\cite{Tanaka2018joint,jiang2018mentornet,han2018co,wang2019symmetric,Kim2019nlnl,yi2019probabilistic,li2020dividemix,Nguyen2020self}. Many works adopt the small-loss trick, which treats examples with small training losses as correctly-labeled. In particular, MentorNet~\cite{jiang2018mentornet} reweights samples according to their loss magnitude so that noisy samples contribute less to the loss. Co-teaching~\cite{han2018co} trains two networks where each network selects small-loss samples in a mini-batch to train the other. DivideMix~\cite{li2020dividemix} fits a Gaussian mixture model on per-sample loss distribution to divide the training data into a clean set and a noisy set. In addition, AUM~\cite{pleiss2020identifying} introduces a margin statistic to identify noisy samples by measuring the average difference between the logit values for a sample's assigned class and its highest non-assigned class. The above methods only consider training datasets that are class-balanced and thus are not applicable for long-tailed label distribution. Recent work~\cite{li2021mopro,li2023stochastic} observes real-world datasets with label noise also have an imbalanced number of samples per-class. Nevertheless, they only inspect a particular case, but we provide a systematic study of noisy labels under various long-tailed scenarios and propose a novel class-wise prototypical noise detection method.

\section{RoLT: Robust Long-Tail Learning with Noisy Labels}\label{sec:method}

\subsection{Problem Setting \& Background}
Given a training dataset $\mathcal{D} = \{\x_i, y_i\}_{i=1}^{N}$, where $\x_i$ is an instance feature vector and $y_i \in \mathcal{C} = [K] = \{1, \dots, K\}$ is the class label assigned to it. The training examples $(\x_i, y_i), 1 \leq i \leq N$ consists of two types: i) a \textit{correctly-labeled example} whose assigned label matches the ground-truth label, i.e., $y_i = y_i^{*}$, where $y_i^{*}$ denotes the ground-truth label of $\x_i$, ii) a \textit{mislabeled example} whose assigned label does not match the ground-truth label, but the input matches one of the classes in $\mathcal{C}$, i.e., $y_i \neq y_i^{*}$ and $y_i^{*} \in \mathcal{C}$. Moreover, the data follows a long-tailed class distribution where the class prior distribution $\mathbb{P}(y)$ is highly skewed, so many underrepresented classes have a very low probability of occurrence. Specifically, we denote the imbalance ratio as $\rho = \max_y \mathbb{P}(y) / \min_y \mathbb{P}(y)$ to indicate the skewness of data. Classes with high $\mathbb{P}(y)$ are referred to as \textit{head classes}, and others are referred to as \textit{tail classes}.

In practice, since the data distribution is unknown, Empirical Risk Minimization (ERM) uses the training data to achieve an empirical estimate of the underlying data distribution. Typically, one minimizes the softmax cross-entropy as following
\begin{equation}
\ell(y, f(\x))=\log \bigg[\sum_{y^{\prime} \in[K]} e^{f_{y^{\prime}}(\x)}\bigg]-f_{y}(\x),
\end{equation}
where $f_y(\x)$ denotes the predictive probability of model $f$ on class $y$. This ubiquitous approach neglects the issue of class imbalance and makes the model biased toward head classes. Moreover, it assumes training examples are correctly-labeled. 

To reduce the impact of noisy labels, sample selection based on the small-loss criterion is one of the most popular approaches. It selects ``easy'' samples for training based on the outputs of the current networks. Specifically, given an arbitrary training example $(\x_i, y_i)$, we can obtain a loss $\ell_{i}$, i.e., $\ell_{i}=\ell\left(f\left(\x_i\right), y_i\right)$. Then a certain proportion of samples with small losses are selected as probably clean samples. The selection can be achieved by a manually set hyperparameter~\cite{han2018co} or a mixture distribution model that fits two-component mixtures on the loss distribution~\cite{arazo2019unsupervised}.

The small-loss criterion was demonstrated to be effective in the existing literature, but we find that it does not work well on the long-tail dataset. It is known that DNNs learn ``easy'' samples first, and on long-tail datasets, DNNs first learn to recognize head classes ahead of tail classes. In light of that, DNNs tend to misclassify tail class examples as head classes. This phenomenon has been widely revealed in recent works~\cite{DBLP:conf/cvpr/ZhouCWC20bbn,DBLP:conf/iclr/KangXRYGFK20}. Therefore, the losses cannot reflect the probability that a sample is mislabeled, especially for tail classes as shown in Figure~\ref{fig:loss_class_0} and~\ref{fig:loss_class_7}.

\subsection{Class-wise Distance-based Sample Selection}
We present a new criterion for selecting noisy labels, called \textit{small-distance}. Formally, we model clean examples of class $k \in [K]$ as if they were distributed around prototype $\c_k \in \mathbb{R}^{D}$, and the likelihood of an example $\x$ belonging to class $k$ decays exponentially with its distance from the prototype $\c_k$, i.e., $\mathbb{P}(\x \mid \c_k) \propto e^{ -dist(\c_k, \x) }$, which is a common assumption about the data distribution~\cite{DBLP:conf/nips/GoldbergerRHS04,DBLP:journals/corr/abs-2104-03066}. Here, $dist(\cdot, \cdot)$ represents any distance measured in the embedding space and is typically set to be the Euclidean distance. Considering the variance of data distribution of different classes, we propose to inspect the distance statistics in a \textit{class-wise} manner.

\begin{table}
\small
\centering
\caption{Average AUC of different classes.}

\begin{tabular}{c c c c c }
\toprule
 Noise level & Many & Medium & Few & All \\
\midrule
{$\gamma=20\%$} & 95.16 & 93.38 & 82.81 &{90.64} \\
{$\gamma=50\%$} & 92.00 & 87.43 & 73.60 & 85.20 \\
\bottomrule
\end{tabular}
\label{tab:auc}
\end{table}

To justify the feasibility of using distance to select clean examples, we compute the AUC based on the distance between examples and their class prototypes for each class separately and report the average value of \textit{Many} (more than 100 images), \textit{Medium} (20$\sim$100 images), \textit{Few} (less than 20 images) and \textit{All} shots in Table~\ref{tab:auc}. The experiment is done on CIFAR-100 with imbalance ratio $\rho=100$, noise level $\gamma=20\%$ and $\gamma=50\%$. It can be seen that the AUC is high even for tail classes, indicating the distance measure may be a useful approach to distinguish clean and noisy examples. Notice that, previous literature~\cite{lee2019rog,wu2020topofilter,Wu_2021_ICCV} also confirms that the representations are resilient to noisy labels.

As mentioned earlier, sample selection can be accomplished using either a thresholding hyperparameter or a mixture distribution model. In line with previous research \cite{arazo2019unsupervised,li2020dividemix}, we adopt the latter approach. 
Specifically, for class $k$, we employ a two-component GMM~\cite{DBLP:journals/pr/PermuterFJ06} to model the distance distribution of clean and noisy samples, i.e., $d \sim \sum_{j=1}^{2} \phi_j \mathcal{N}(\mu_j, \sigma_j^2)$, where $d = dist(\c_k, \x), \forall \x \in \mathcal{D}_k$ and $\phi_j$ denotes weight of the $j$-th component. Note that we have $\sum_{j=1}^2 \phi_j = 1$. Without loss of generality, we assume $\mu_1 < \mu_2$. Since clean examples are located around the prototype while noisy examples spread out, we flag $\x$ as clean if and only if $\mathbb{P}(d \mid \mu_1, \sigma_1) > \mathbb{P}(d \mid \mu_2, \sigma_2)$.
We thus perform noise detection by estimating the Gaussians' parameters from distance statistics.
In particular, for each class $k$, we compute its prototype as the normalized average of the embeddings for training examples by
\begin{equation}\label{equ:compute-prototype}
\boldsymbol{c}_{k} \leftarrow \operatorname{Normalize}\bigg( \frac{1}{|\mathcal{D}_k|} \sum_{\x_i \in \mathcal{D}_k } f_\theta (\x_i) \bigg), \; \forall k\in [K]
\end{equation}
where $f_\theta(\x)$ denotes the extracted feature of $\x$ and $\forall \x_i \in \mathcal{D}_k$. Given $\c_k$, the distances between $\c_k$ and examples of class $k$ are obtained by 
\begin{equation}\label{equ:distance}
    dist(\c_k, \x_i) = \big\|\c_k - f_\theta(\x_i) \big\|_2^2,
\end{equation}
We then fit a two-component Gaussian mixture model to maximize the log-likelihood value, i.e., $\max \sum_{i=1}^{|\DD_k|} \log (\sum_{j=1}^2 \phi_j \mathbb{P}(d_k(\x_i) \mid \mu_j, \sigma_j))$, where $d_k(\x_i) = dist(\c_k, \x_i)$ for $\x_i \in \mathcal{D}_k$.

For simplicity, we denote the clean (noisy) data of class $k$ as $\XX_k$ ($\SS_k$). Note that we have $\DD_k = \XX_k \bigcup \SS_k$. Therefore, we obtain a subset of clean examples by $\mathcal{X} = \bigcup_{k=1}^{K} \XX_k$ and noisy examples by $\mathcal{S} = \bigcup_{k=1}^{K} \SS_k$. 
It is also verified that the Gaussian mixture model can be used to distinguish clean and noisy data because of its flexibility in the sharpness of distribution in previous literature~\cite{li2020dividemix}. We also observe that the proposed method works well on real-world datasets where the Gaussian distribution assumption is not perfectly satisfied. 
Recall that $\DD_k$ may contain noisy labels, the estimate of $\c_k$ in~\eqref{equ:compute-prototype} is inaccurate and the split of $\DD_k = \XX_k \bigcup \SS_k$ is problematic. To remedy this, we refine class prototypes using $\XX_k$ rather than $\DD_k$, and acquire a new split of $\DD_k$. By doing this, $\XX_k$ retains most of correctly-labeled examples of class $k$ as well as fewer mislabeled examples. 

\subsection{Soft Pseudo-Labeling with Label Distribution}
For each example that is likely to be mislabeled, its original discrete noisy label is transformed into a continuous label distribution to incorporate the uncertainty of the ground-truth label. The benefits of using label distributions are two-fold. First, it mitigates the influence of noisy labels~\cite{lukasik2020does}; Second, it compensates for the learning of data scarcity tail classes~\cite{DBLP:journals/corr/abs-2104-00466}.
To this end, the underrepresented tail classes will receive significant improvements, which is crucial in LTL.
To generate label distributions, a direct approach is to leverage the prediction of the ERM (Empirical Risk Minimization) model. However, the ERM is known to be biased toward head classes~\cite{DBLP:journals/corr/abs-2104-00466}. Hence, refining noisy labels using the predictions of ERM may be sub-optimal for examples of tail classes. In contrast, the NCM (Nearest Class Mean) classifier can yield a balanced classification boundary~\cite{DBLP:conf/iclr/KangXRYGFK20}. Specifically, we find that the NCM classifier produces much higher recall on tail classes than the ERM in experiments. By aggregating the predictive information from the ERM and NCM classifiers, we construct diverse soft pseudo-labels for detected noisy examples. To amend the misflag of the noise detector, we also take into account the original labels as a source of soft pseudo-labels. Since the predictions of ERM and NCM classifiers may disagree, we further remedy this by the label smoothing technique~\cite{DBLP:journals/corr/abs-2104-00466}.

\begin{figure}[!h]
    \centering
\includegraphics[width=1\linewidth]{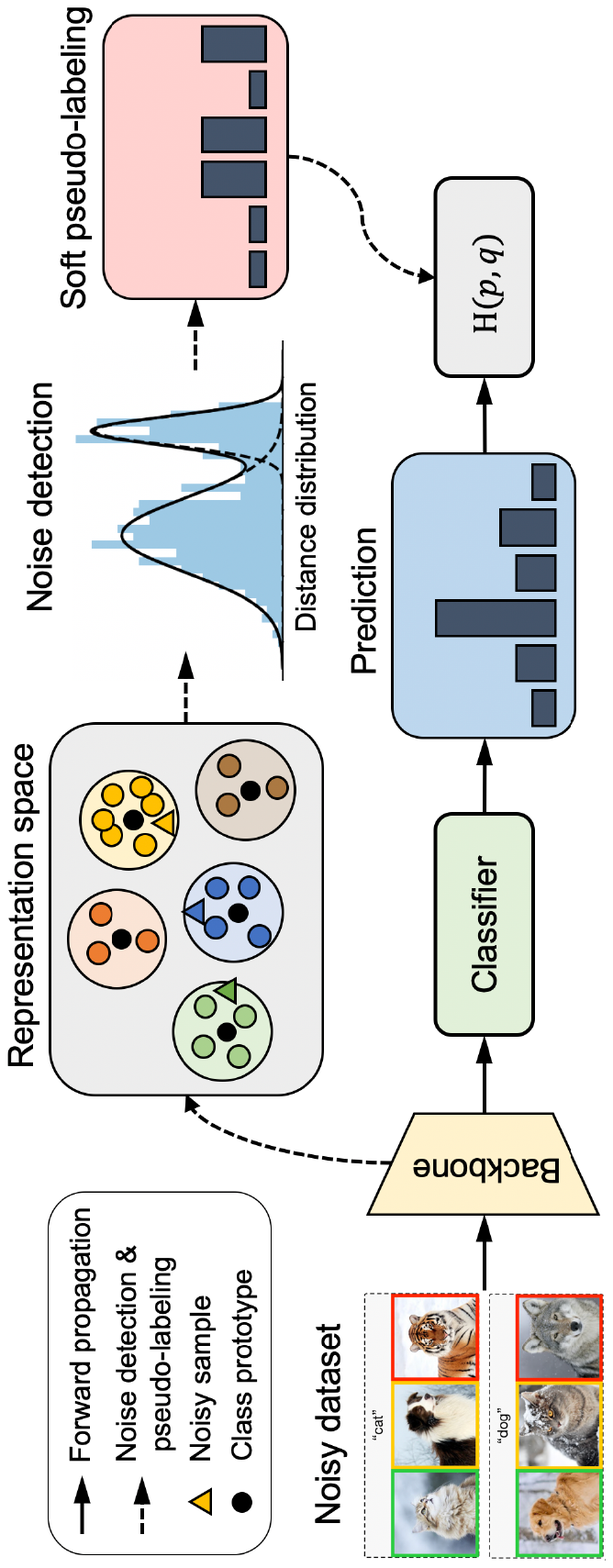}
\caption{The proposed framework~\algo.}\label{fig:framwork}
\end{figure}

Given the predictions $\hat{y}^{erm} = \arg \max_k f(\x)$, $\hat{y}^{ncm} = \arg \min_k \left\|\c_k - f_{\theta}(\x) \right\|_2$, and original noisy label $y$, we construct the guessing label set $\mathcal{G} = \{\hat{y}^{erm}, \hat{y}^{ncm}, y\}$ and generate the label distribution $\tilde{\y} \in \mathbb{R}^{K}$ for $\x$. For $k \in [K]$, we compute
\begin{equation}\label{equ:soft-label}
\tilde{y}_{k} = \left\{\begin{array}{ll}
\frac{1}{4} \sum_{\hat{y} \in \mathcal{G}} \mathbb{I}(\hat{y} = k) + \frac{1}{4K}& \text { if } k \in \mathcal{G} \\[0.2cm]
\frac{1}{4K} & \text { otherwise. }
\end{array}\right.
\end{equation}
Here, $\mathbb{I}(\cdot)$ is an indicator which returns $1$ if the condition is true, otherwise $0$. Considering the classification task with four classes (i.e., $K=4$) and $\mathcal{G} = \{1, 4, 2\}$, the soft pseudo-label would be $\tilde{\y} = [\frac{5}{16}, \frac{5}{16}, \frac{1}{16}, \frac{5}{16}]$.
The targets $\hat{y}^{erm}$ and $\hat{y}^{ncm}$ can be set equal to the model output, but using a running average is more effective which is known as temporal ensembling~\cite{laine2016temporal} in semi-supervised learning. For ERM or NCM classifier, let $\boldsymbol{z}_i(t) \in \mathbb{R}^{K}$ be the output logits vector (pre-softmax output) for example $\x_i$ at iteration $t$ of training, we update the momentum logits by
\begin{equation}
\boldsymbol q_{i}(t)=\alpha \boldsymbol q_{i}(t-1)+(1-\alpha) \boldsymbol{z}_{i}(t),
\end{equation}
where $0 \leq \alpha < 1$ is the combination weight. For each iteration $t$, we then obtain $\hat{y}^{erm}$ and $\hat{y}^{ncm}$ using softmax of $\boldsymbol q_{i}(t)$. Having acquired $\XX$, $\SS$, and soft pseudo-labels, we compute the cross-entropy loss for clean examples using original training labels by
\begin{equation}
    \LL_{\XX} = \frac{1}{|\XX|} \sum_{\x_i \in \XX} H (\y_i, f(\x_i)),
\end{equation}
where $\y_i$ is the one-hot label vector for $\x_i$. For noisy examples, the loss function is computed by
\begin{equation}
    \LL_{\SS} = \frac{1}{|\SS|} \sum_{\x_i \in \SS} H (\tilde{\y}_i, f(\x_i)),
\end{equation}
where $H(\boldsymbol{q}, \boldsymbol{p}) = -\sum^{K}_{k=1} q_k \log \big( \frac{\exp p_k}{ \sum_{j=1}^{K} \exp p_j } \big) $ is the cross-entropy between distributions $\boldsymbol{q}$ and $\boldsymbol{p}$. Overall, the training objective is $\LL = \LL_{\XX} + \LL_{\SS}$. 
We illustrate the whole proposed framework in \Cref{fig:framwork}.

\subsection{Combining with Semi-Supervised Learning}
The proposed method can be further improved by using a well-established semi-supervised learning approach, where clean and noisy examples are viewed as labeled and unlabeled data, respectively. Inspired by DivideMix~\cite{li2020dividemix} and ELR+~\cite{liu2020early}, we use two separate neural networks, where the target of each network is computed from the output of the other network. For fair comparisons, we replace the sample selection module of DivideMix with our proposed class-wise prototypical noise detector. We call this improved method \algo+.


\begin{table*}[!h]
\small
\centering
\caption{Comparisons of methods using a \textbf{single model}. Test accuracy (\%) on \textit{CIFAR} datasets with different imbalance ratios and noise levels. }\label{tab:cifar}
\setlength{\tabcolsep}{4pt}
\begin{threeparttable}[!h]
\begin{tabular}{ l c | c c c | c c c | c c c | c c c }
\toprule
 &  & \multicolumn{6}{ c| }{CIFAR-10} & \multicolumn{6}{ c }{CIFAR-100} \\
\midrule
\multicolumn{2}{ l| }{Noise level} & \multicolumn{3}{ c| }{20\%} & \multicolumn{3}{ c| }{50\%} & \multicolumn{3}{ c| }{20\%} & \multicolumn{3}{ c }{50\%}\\
\midrule
\multicolumn{2}{ l| }{Imbalance ratio} & 10 & 50 & 100  & 10 & 50 & 100  & 10 & 50 & 100 & 10 & 50 & 100 \\
\midrule
\multirow{2}{*}{CE}  & Best  & 75.61 & 63.60 & 62.17  & 63.25 & 43.38 & 38.71  & 43.27 & 30.27 & 26.21  & 26.92 & 16.97 & 14.23 \\
 & Last  & 73.86 & 58.60 & 54.15  & 48.39 & 32.73 & 27.05  & 43.06 & 30.04 & 26.08  & 26.60 & 15.59 & 13.47 \\
\midrule
\multirow{2}{*}{LDAM}  & Best  & 82.37 & 71.34 & 66.26  & 60.30 & 42.95 & 36.66  & 48.14 & 33.43 & 29.70  & 29.62 & 17.51 & 14.19 \\
 & Last  & 82.07 & 71.11 & 65.88  & 59.10 & 38.33 & 33.38  & 47.89 & 33.30 & 29.50  & 29.38 & 17.29 & 13.24 \\
\midrule
\multirow{2}{*}{LDAM-DRW}  & Best  & 83.73 & 76.41 & 72.28  & 67.93 & 48.88 & 43.23  & 50.44 & 36.60 & 32.27  & 32.24 & 19.48 & 15.21 \\
 & Last  & 83.67 & 75.67 & 71.08  & 67.68 & 47.38 & 41.45  & 50.29 & 36.16 & 32.05  & 31.72 & 19.23 & 14.75 \\
\midrule
\multirow{2}{*}{BBN}  & Best  & 77.81 & 68.01 & 64.51  & 64.71 & 46.22 & 36.72  & 47.60 & 31.07 & 28.79  & 30.01 & 19.75 & 14.56 \\
 & Last  & 76.81 & 67.48 & 64.24  & 53.76 & 43.35 & 34.83  & 47.26 & 30.76 & 28.56  & 29.42 & 19.55 & 14.34 \\
\midrule
\multirow{2}{*}{cRT}  & Best  & 76.15 & 65.02 & 59.92  & 64.15 & 43.26 & 36.73  & 42.56 & 30.23 & 26.31  & 25.55 & 17.47 & 16.01 \\
 & Last  & 75.05 & 64.22 & 58.47  & 62.75 & 41.87 & 34.55  & 41.56 & 30.08 & 26.18  & 23.94 & 17.34 & 15.94 \\
\midrule
\multirow{2}{*}{MW-Net}  & Best  & 82.19 & 71.63 & 67.26 & 72.12 & 56.09 &  46.36  & 50.20 & 36.68 & 31.77 & 37.50 & 23.99 &\bf 21.24  \\
 & Last  & 77.67 & 64.12 & 58.23  & 59.68 & 45.39 & 37.05  & 47.82 & 34.45 &  29.57 & 33.14 & 20.33 & 18.82  \\
\midrule
\multirow{2}{*}{HAR-DRW}  & Best  & 82.43 & 67.44 & 67.88  & 67.39 & 52.35 & 42.80  & 46.24 & 28.86 & 26.29  & 31.30 & 16.75 & 14.78 \\
 & Last  & 78.44 & 61.08 & 62.73  & 64.75 & 45.06 & 40.07  & 43.04 & 26.11 & 24.71  & 26.96 & 13.87 & 12.42 \\
\midrule
& Best  &\bf 85.04 &\bf 77.86 &\bf 73.84  &\bf 77.11 &\bf 60.15 &\bf 55.32  &\bf 53.41 &\bf 38.94 &\bf 33.36  &\bf 39.22 &\bf 25.51 & 20.61 \\
\multirow{-2}{*}{{\algo}}
& Last  &\bf 84.95 &\bf 77.65 &\bf 73.54  &\bf 76.94 &\bf 59.59 &\bf 54.55  &\bf 53.22 &\bf 38.77 &\bf 33.20  &\bf 39.01 &\bf 25.35 &\bf 20.45 \\
\bottomrule
\end{tabular}
%
%

\end{threeparttable}
\end{table*}

\begin{table*}[h]
\small
\centering
\caption{Test accuracy (\%) of methods using \textbf{two models} on CIFAR datasets.}\label{tab:dividemix}
\setlength{\tabcolsep}{4.3pt}
\begin{threeparttable}[!h]
\begin{tabular}{ l c | c c c | c c c | c c c | c c c }
\toprule
 &  & \multicolumn{6}{ c| }{CIFAR-10} & \multicolumn{6}{ c }{CIFAR-100} \\
\midrule
\multicolumn{2}{ l| }{Noise level} & \multicolumn{3}{ c| }{20\%} & \multicolumn{3}{ c| }{50\%} & \multicolumn{3}{ c| }{20\%} & \multicolumn{3}{ c }{50\%}\\
\midrule
\multicolumn{2}{ l| }{Imbalance ratio} & 10 & 50 & 100  & 10 & 50 & 100  & 10 & 50 & 100 & 10 & 50 & 100 \\
\midrule
\multirow{2}{*}{ELR+}  & Best  & \bf 88.96 & \bf 80.21 & 69.60  & 85.02 & 56.96 & 48.72  & 54.01 & 49.64 & 38.40  & 49.53 & 30.12 & 21.58 \\
 & Last  & 88.09 & \bf 79.69 & 66.67  & 84.08 & 48.14 & 43.11  & 53.32 & 48.37 & 38.12  & 49.06 & 29.68 & 20.47 \\
\midrule
\multirow{2}{*}{DivideMix}  & Best  & 88.79 & 75.34 & 66.90  & 87.54 & 67.92 & 61.81  & 63.79 & 49.64 & 43.91  & 49.35 & 36.52 & 31.82 \\
 & Last  &\bf 88.10 & 73.48 & 63.76  & 86.88 & 65.22 & 59.65  & 63.17 & 48.37 & 42.59  & 48.87 & 35.72 & 31.05 \\
\midrule
 & Best  & 87.95 & 77.26 &\bf 72.31  &\bf 88.17 &\bf 75.11 &\bf 64.42  &\bf 64.22 &\bf 51.01 &\bf 45.35  &\bf 53.31 &\bf 39.78 &\bf 35.29 \\
\multirow{-2}{*}{{\algo+}} 
 & Last  & 87.54 & 75.90 &\bf 69.12  &\bf 87.45 &\bf 73.92 &\bf 61.15  &\bf 63.31 &\bf 49.40 &\bf 43.16  &\bf 52.44 &\bf 39.27 &\bf 34.43 \\
\bottomrule
\end{tabular}

\end{threeparttable}
\end{table*}

\subsection{Difference with Prior Works}
Class prototypes are employed in some previous literature and we discuss the differences between this paper and some related works.
In few-shot learning, prototypical networks\cite{DBLP:conf/nips/SnellSZ17} learn a metric space in which classification can be performed by computing distances to prototype representations of each class. Compared to other methods, prototypical networks reflect a simpler inductive bias that is beneficial in this limited-data regime, and achieve excellent results.
In long-tailed recognition, OLTR\cite{DBLP:conf/cvpr/0002MZWGY19} proposes to use the distances between samples and prototypes to handle open-set recognition.
In self-supervised learning, PCL\cite{DBLP:conf/iclr/0001ZXH21} proposes the ProtoNCE loss which encourages representations to be closer to their assigned prototypes.
However, the above-mentioned works do not consider using class prototypes to detect noisy labels in long-tailed class distribution. Moreover, the distances calculated between examples and their class prototypes are utilized in a \textit{class-wise} manner to mitigate the influence of long-tailed class distribution.

\section{Experiments}

\subsection{Results on Simulated Datasets}
\textbf{Setting.} We test~\algo~on CIFAR-10 and CIFAR-100 under various imbalanced ratio $\rho$ and noise level $\gamma$. CIFAR-10 consists of 50k images across 10 classes, while CIFAR-100 contains 50k images but with 100 classes. For each dataset, we first simulate the long-tailed dataset following the same setting as LDAM~\cite{DBLP:conf/nips/CaoWGAM19ldam}. The long-tailed imbalance follows an exponential decay in sample sizes across different classes. To inject noisy labels, we define the noise transition matrix or the asymmetric noise policy to construct the training set. In particular, we consider imbalance ratio to be $\rho \in \{10, 50, 100\}$ and noise level to be $\gamma \in \{20\%, 50\%\}$.

\textbf{Results under simulated noise.} Table~\ref{tab:cifar} and Table~\ref{tab:dividemix} summarize the results for CIFAR-10 and CIFAR-100. As shown in the results, previous LTL methods (i.e., LDAM, BBN~\cite{DBLP:conf/cvpr/ZhouCWC20bbn}, cRT~\cite{DBLP:conf/iclr/KangXRYGFK20}) dreadfully degrade their performance as the noise level and imbalance ratio increase, while our methods retain robust performance. In particular, compared with CE,~\algo~improves the test accuracy by 8\% on average. It can be observed that the improvement becomes more significant at high noise levels, benefiting from proposed noise detection and soft pseudo-labeling. Further application of Deferred Re-Weighting (DRW)~\cite{DBLP:conf/nips/CaoWGAM19ldam} enhances the performance by favoring the tail classes. This clearly demonstrates the importance of correcting noisy labels in the training data. 
Notice that MW-Net~\cite{DBLP:conf/nips/ShuXY0ZXM19} and HAR-DRW~\cite{cao2021heteroskedastic} are proposed to handle label noise and class imbalance. Our method consistently outperforms them by a large margin.

We further compare~\algo+ with DivideMix~\cite{li2020dividemix} and ELR+~\cite{liu2020early}, which are the most popular methods for learning with noisy labels. The results are given in Table~\ref{tab:dividemix}. First, we can see that the performance of ELR+ significantly drops as the training set becomes class-imbalanced. DivideMix is relatively robust to class imbalance than ELR+ by imposing the uniform predictions regularization in its objective. In contrast, our method~\algo+ achieves performance improvements in test accuracy by 2.57\% on average. This validates the superiority of our noise detector. 

\begin{table}
\small
\centering
\caption{ Results on \textit{CIFAR-10} with asymmetric noise. }\label{tab:asym}

\setlength{\tabcolsep}{3pt}
\begin{tabular}{ l c | c c c | c c c }
\toprule
\multicolumn{2}{ l| }{Noise level} & \multicolumn{3}{ c| }{20\%} & \multicolumn{3}{ c }{40\%} \\
\midrule
\multicolumn{2}{ l| }{Imbalance ratio} & 10 & 50 & 100  & 10 & 50 & 100 \\
\midrule
\multirow{2}{*}{DivideMix}  & Best  & 83.48 & 73.13 & 66.51  & 78.85 & 67.74 & 59.63   \\
 & Last  & 78.77 & 56.90 & 44.50  & 74.50 & 46.48 & 32.68   \\
\midrule
  & Best  &\bf 84.05 &\bf 75.93 &\bf 68.53  &\bf 79.98 &\bf 71.44 &\bf 65.56   \\
\multirow{-2}{*}{{\algo+}}
 & Last  &\bf 79.81 &\bf 67.13 &\bf 60.38  &\bf 74.68 &\bf 60.07 &\bf 54.05  \\
\bottomrule
\end{tabular}
\end{table}

\textbf{Results under asymmetric noise.} We further verify the effectiveness of the proposed detection method and the robust framework under asymmetric label noise. We conduct experiments on the long-tail CIFAR-10 dataset using the noise injection policy following previous works~\cite{li2020dividemix,liu2020early}. From \Cref{tab:asym}, it can be seen that \algo+ consistently outperforms DivideMix in all cases. It is interesting to observe that the performance gap between \textit{best} and \textit{last} widens as the imbalance ratio becomes large. This reflects that the model is easy to collapse under asymmetric noise and class imbalance. The proposed method can alleviate this issue substantially.


\subsection{Results on Real-World Dataset}
\paragraph{Results on WebVision dataset.}
We test the performance of our method on a real-world dataset. WebVision~\cite{webvision} contains 2.4 million images collected from Flickr and Google, characterized by real-world label noise and class imbalance. The estimated noise level is 20\%. Following previous literature, we use a subset of WebVision, known as mini WebVision, which includes the first 50 classes. In Table~\ref{exp:webvision}, we compare \algo\ with several previous approaches, including D2L~\cite{ma2018dimensionality}, MentorNet~\cite{jiang2018mentornet}, Co-teaching~\cite{han2018co}, Iterative-CV~\cite{chen2019understanding}, HAR~\cite{cao2021heteroskedastic}, DivideMix~\cite{li2020dividemix}, and ELR+~\cite{liu2020early}. \algo\ achieves superior performance than DivideMix and ELR+, particularly in terms of top-5 accuracy.

\begin{table*}[!h]
\centering
\caption{Accuracy (\%) on mini \textit{WebVision} and \textit{ImageNet} validation sets.}\label{exp:webvision}

\setlength{\tabcolsep}{4pt}
\begin{tabular}{l c c c c c c c c}
\toprule
& Metric & D2L & MentorNet & Co-teaching & Iterative-CV & HAR & DivideMix & {\algo+} \\
\midrule
\multirow{2}{*}{Webvision} & top1 & 62.68 & 63.00 & 63.58 & 65.24 & 75.5 & 77.32 & \textbf{77.64} \\
                           & top5 & 84.00 & 81.40 & 85.20 & 85.34 & 90.7 & 91.64 & \textbf{92.44}\\
\midrule
\rule{0pt}{2.2ex} \multirow{2}{*}{ImageNet} & top1 & 57.80 & 57.80 & 61.48 & 61.60 & 70.3 & \textbf{75.20} & 74.64\\
                           & top5 & 81.36 & 79.92 & 84.70 & 84.98 & 90.0 & 90.84 & \textbf{92.48}\\
\bottomrule
\end{tabular}

\end{table*}

To further uncover the advantages of our method, we run experiments by controlling the imbalance ratio of Webvision dataset. The test accuracy is reported in the Table~\ref{tab:webvision-control}. From the results, we can see that the superiority of our method is more significant as the imbalance ratio increases.

\begin{table*}[!h]
\centering
\caption{Top1 (top5) accuracy on \textit{Webvision} and \textit{ImageNet}. }\label{tab:webvision-control}
\label{tab:ablation}
\setlength{\tabcolsep}{0.1pt}
\begin{tabular*}{0.9\textwidth}{@{\extracolsep{\fill}}*{6}{l}}
\toprule
\multicolumn{3}{c}{Imbalance ratio $\rho=50$} &
\multicolumn{3}{c}{Imbalance ratio $\rho=100$} \\
\cmidrule{1-3}\cmidrule{4-6}
Method & Webvision  &  ImageNet &
Method & Webvision  &  ImageNet \\
\cmidrule{1-3}\cmidrule{4-6}
DivideMix & 64.56	(83.56)	& 62.68	(85.24) &
DivideMix & 55.76 (73.48) & 53.92 (74.00) \\
w/ DRW    & 68.16 (84.92) & 66.12 (85.40) &
w/ DRW    & 60.28  (74.60) & 59.04 (75.68) \\
\cmidrule{1-3}\cmidrule{4-6}
RoLT+     & 66.28	(88.68) & 64.76	(89.96) &
RoLT+     & 60.68 (87.84) & 59.68 (88.52) \\
w/ DRW    & 70.08 (88.52) & 67.28 (90.12) &
w/ DRW    & 65.48 (87.32) & 64.80 (87.08) \\   
\bottomrule
\end{tabular*}

\end{table*}

\paragraph{Results on ImageNet-LT dataset.}

\begin{table}
\centering
\small
\caption{Accuracy (\%) on \textit{ImageNet-LT} dataset.}
\label{tab:imagenet}
\setlength{\tabcolsep}{4pt}

\begin{tabular}{c c c c c} 
\toprule
& Method & - & w/ cRT & w/ LA  \\
\midrule 
 \multirow{3}{*}{$\gamma=20\%$} & CE & 28.18 & 34.15 & 34.06 \\
& ELR & 26.58 & 35.21 & 34.05 \\
& RoLT & \textbf{29.57} & \textbf{35.76} &  \textbf{35.09}   \\
\midrule
  \multirow{3}{*}{$\gamma=50\%$} & CE & 17.80 & 21.85 & 22.71\\
  & ELR & 17.33 & 22.80 & 22.60\\
 & RoLT &  \textbf{21.53} &  \textbf{25.61} &  \textbf{25.50}\\
\bottomrule
\end{tabular}

\end{table}
We compare our method with baselines (CE and ELR) on a large long-tail benchmark, i.e., \textit{ImageNet-LT}, by combining with methods and strategies for class-imbalanced datasets, i.e. Classifier Re-training (cRT)~\cite{DBLP:conf/iclr/KangXRYGFK20} and Logit Adjustment (LA)~\cite{menon2021logitadjust}. The ImageNet-LT dataset consists of 1000 classes and its imbalance ratio is 200. We simulate various levels of label noise to assess the performance under different noise levels. From the results in \Cref{tab:imagenet}, we can clearly see that our method outperforms other methods, particularly under high noise levels.


\begin{figure}[!b]
    \centering
    \begin{subfigure}[b]{0.22\textwidth}
        \centering
        \includegraphics[width=\textwidth]{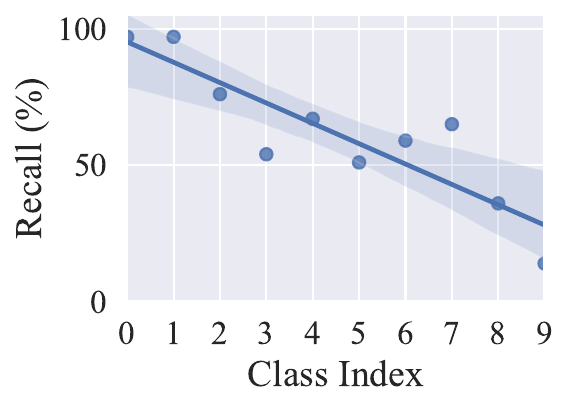}
        \caption{ERM on \textit{CIFAR-10}}
        \label{fig:ce_recall_cifar10}
    \end{subfigure}
    \begin{subfigure}[b]{0.22\textwidth}
        \centering
        \includegraphics[width=\textwidth]{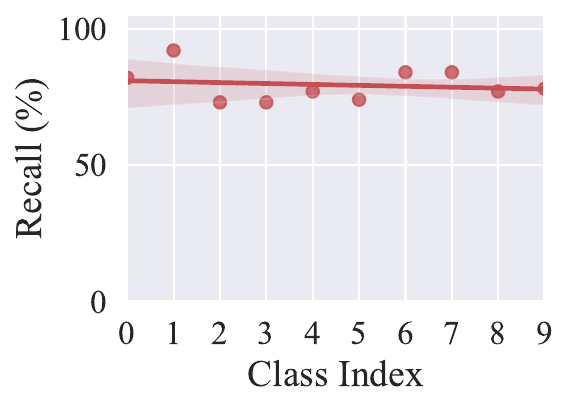} 
        \caption{NCM on \textit{CIFAR-10}}\label{fig:ncm_recall_cifar10}
    \end{subfigure}
    \begin{subfigure}[b]{0.22\textwidth}
        \centering
        \includegraphics[width=\textwidth]{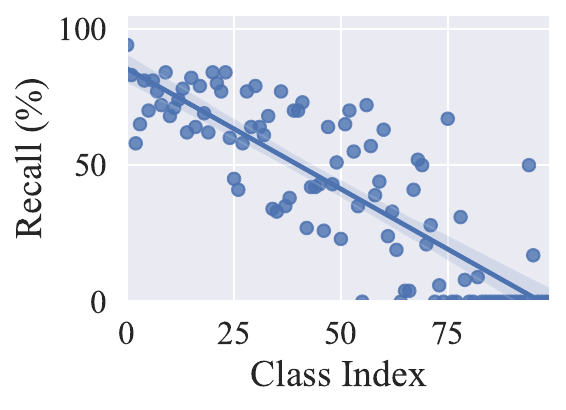} 
        \caption{ERM on \textit{CIFAR-100}}\label{fig:ce_recall_cifar100}
    \end{subfigure}
    \begin{subfigure}[b]{0.22\textwidth}
        \centering
        \includegraphics[width=\textwidth]{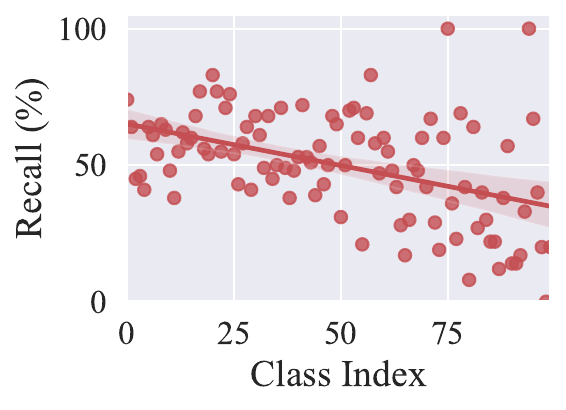} 
        \caption{NCM on \textit{CIFAR-100}}\label{fig:ncm_recall_cifar100}
    \end{subfigure}
    \caption{Per-class recall of ERM and NCM classifiers.} 
    \label{fig:recall-ce-ncm}
\end{figure}

\begin{figure}[h]
    \centering
    \includegraphics[width=0.4\textwidth]{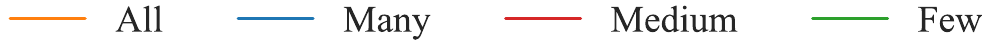}\\
    \begin{subfigure}[b]{0.22\textwidth}
        \centering
        \includegraphics[width=\textwidth]{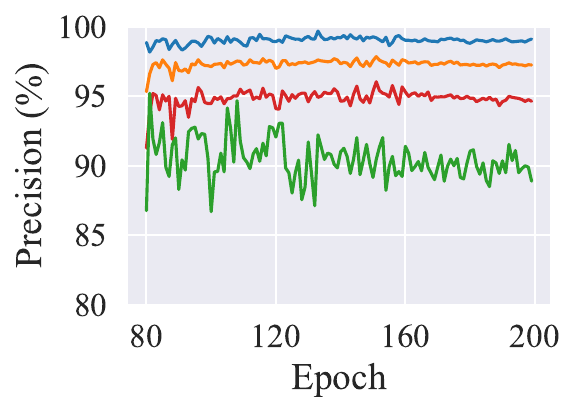}
        \caption{ \textit{CIFAR-10} Precision}
        \label{fig:gmm_precision_cifar10}
    \end{subfigure}
    \begin{subfigure}[b]{0.22\textwidth}
        \centering
        \includegraphics[width=\textwidth]{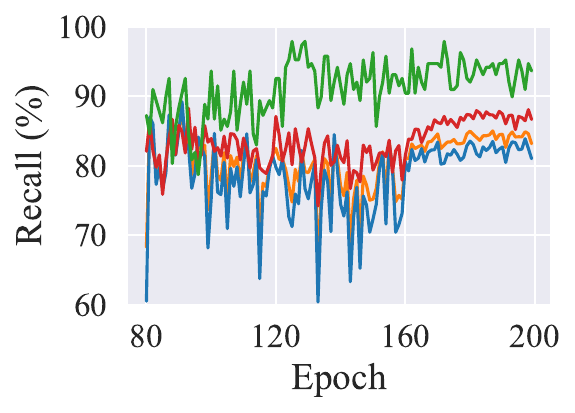} 
        \caption{\textit{CIFAR-10} Recall}\label{fig:gmm_recall_cifar10}
    \end{subfigure}
    \begin{subfigure}[b]{0.22\textwidth}
        \centering
        \includegraphics[width=\textwidth]{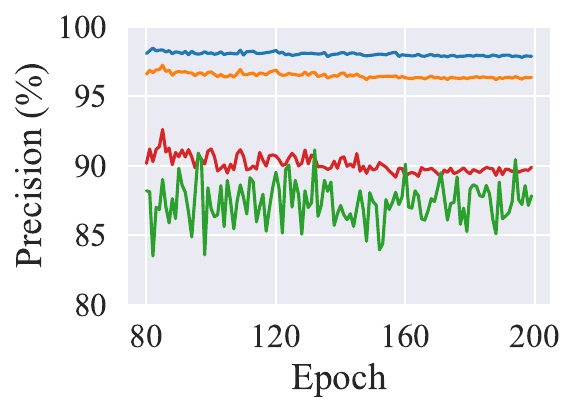} 
        \caption{\textit{CIFAR-100} Precision}\label{fig:gmm_precision_cifar100}
    \end{subfigure}
    \begin{subfigure}[b]{0.22\textwidth}
        \centering
        \includegraphics[width=\textwidth]{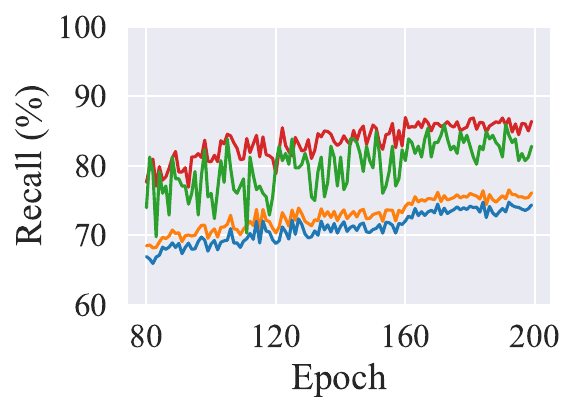} 
        \caption{\textit{CIFAR-100} Recall}\label{fig:gmm_recall_cifar100}
    \end{subfigure}
    \caption{Precision and Recall of selected samples.}
    \label{fig:gmm-precision-recall}
\end{figure}

\begin{table}[!h]
\footnotesize
\centering
\caption{ Ablation studies on pseudo-labeling. Test accuracy on \textit{CIFAR-100} is reported.}\label{tab:pseudo-label}
\setlength{\tabcolsep}{6pt}
\begin{tabular}{ c  c  c c c c }
\toprule
\multirow{2}{*}{DRW} & \multirow{2}{*}{ Method }  & \multicolumn{4}{ c}{$\gamma=50\%$}  \\
 &  & Many & Med. & Few & All  \\
\midrule
\xmark  & Noisy    & 32.06 & 7.89 & 0.04 & 14.23 \\
\xmark & ERM    & 38.83 & 12.05 & 0.89 & 18.41 \\
\xmark & Soft (w/o LS)    & 36.03 & 15.11 & 2.19 & 18.94 \\
\xmark & Soft (w/ LS)    & 37.20 & 15.97 & 1.74 &\bf 19.56 \\
\midrule
\rule{0pt}{2.2ex}\cmark  & Noisy    & 23.77 & 14.53 & 3.41 & 14.76 \\
\cmark  & ERM  & 32.80 & 17.05 & 2.30 & 18.58 \\
\cmark  & Soft (w/o LS)   & 30.31 & 18.92 & 4.93 & 19.13  \\
\cmark  & Soft (w/ LS)  & 30.94 & 21.32 & 6.22 &\bf 20.61 \\
\bottomrule
\end{tabular}

\end{table}

\subsection{Further Analysis and Ablation Studies}

\textbf{Efficacy of the noise detector.}
To further support our motivation, we compare the performance of the ERM and NCM classifiers in Figure~\ref{fig:recall-ce-ncm}. It can be seen that NCM produces more balanced recall across classes, while ERM tends to predict examples as head classes, resulting in low recall for tail classes. Figure~\ref{fig:gmm-precision-recall} shows the precision and recall of selected clean examples by our method. To better understand~\algo, we construct three groups of classes for CIFAR-100 by: many (more than 100 images), medium (20$\sim$100 images), and few (less than 20 images) shots; and CIFAR-10 by: many ($\{0,1\}$), medium($\{2,\dots,6\}$), and few ($\{7,8,9\}$) shots according to class indices.~\algo~maintains high precision and recall, which validates the effectiveness of our method. This experiment is conducted under the imbalance ratio $\rho=100$ and noise level $\gamma =30\%$.

\textbf{Efficacy of the soft pseudo-labeling.}
We investigate the effectiveness of soft pseudo-labeling by comparing it with three other methods: (i) ``\textit{Noisy}'' keeps the noisy labels, (ii) ``\textit{ERM}'' rectifies labels via the ERM predictions, (iii) ``\textit{Soft (w/o LS)}'' uses the label distribution without label smoothing (w/o LS). The ``\textit{Soft (w/o LS)}'' method produces label distribution as following:
\begin{equation}
\tilde{y}_{k} = \left\{\begin{array}{ll}
\frac{1}{3} \sum_{\hat{y} \in \mathcal{G}} \mathbb{I}(\hat{y} = k) & \text { if } k \in \mathcal{G} \\[0.2cm]
0 & \text { otherwise. }
\end{array}\right.
\end{equation}
We report the results in Table~\ref{tab:pseudo-label} for noise level $\gamma =50\%$ and imbalance ratio $\rho = 100$. We observe that ERM and soft pseudo-labeling significantly improve the performance by over 4\% in test accuracy, and the improvement is more significant under high noise levels. Moreover, the soft pseudo-labeling outperforms its ERM and ``w/o LS'' counterpart in most cases, demonstrating that label smoothing and label guessing can provide diverse and informative supervision under imperfect training labels. 

\textbf{Results on class-balanced datasets.}
We compare the performance of our method with DivideMix on balanced datasets with noise level $\rho \in \{20\%, 50\%\}$. The results are reported in Table~\ref{tab:cifar_balanced} and our method is comparable with DivideMix. This shows that the proposed prototypical noise detector also works well on balanced datasets.

\begin{table}[!h]
\centering
\caption{ Test accuracy (\%) on class-balanced datasets.}\label{tab:cifar_balanced}
\begin{tabular}{ l c | c | c | c | c }
\toprule
 &  & \multicolumn{2}{ c| }{CIFAR-10} & \multicolumn{2}{ c }{CIFAR-100} \\
\midrule
\multicolumn{2}{ l| }{Noise level} & 20\% & 50\%  & 20\% & 50\%  \\
\midrule
\multirow{2}{*}{DivideMix}  & Best  &\bf 92.79 &\bf 95.03  & 77.25 & 73.84 \\
 & Last  &\bf 92.41 &\bf 94.63  & 77.03 &\bf 73.42 \\
\hline
\multirow{2}{*}{\textbf{\algo+}}  & Best  & 92.46 & 94.59  &\bf 78.60 &\bf 74.11 \\
 & Last  & 92.01 & 94.41  &\bf 78.14 & 73.35 \\
\bottomrule
\end{tabular}
\end{table}

\begin{table}[!h]
\setlength{\tabcolsep}{4pt}
\centering
\caption{ Test accuracy (\%) on clean datasets. }\label{tab:cifar_clean}

\begin{threeparttable}[!h]
\begin{tabular}{ l | c c c | c c c }
\toprule
 & \multicolumn{3}{ c| }{CIFAR-10} & \multicolumn{3}{ c }{CIFAR-100} \\
\midrule
Imb. ratio  & 10 & 50 & 100  & 10 & 50 & 100 \\
\midrule
CE  & 86.75 & 77.38 & 71.83  & 56.31 & 44.15 & 38.88  \\
LDAM  & 86.38 & 77.62 & 74.31  & 55.66 & 43.61 & 39.25  \\
LDAM-DRW  & 87.29 & 81.25 & 78.78  & 57.21 & \bf 47.30 & \bf 42.93  \\
BBN  & 87.83 & 81.19 & 78.87  & 58.08 & 45.62 & 40.09  \\
cRT  & 86.78 & 77.30 & 71.18  & 56.62 & 43.01 & 39.44  \\
MW-Net$^\dagger$  &\bf 87.99 & 79.58 & 74.92  &\bf 58.66 & 46.72 & 42.10 \\
HAR-DRW  & 87.81 & 79.82 & 75.99  & 56.89 & 43.34 & 40.78  \\
\textbf{\algo}  &\bf 87.99 & 80.50 & 77.70  & 57.47 & 45.38 & 39.35  \\
\textbf{\algo-DRW}  & 87.75 &\bf 83.02 & \bf 80.57  & 57.48 & 47.21 & 41.70  \\
\bottomrule
\end{tabular}
\begin{tablenotes}
     \item[$\dagger$] MW-Net uses a $1k$ clean and class-balanced validation set.
\end{tablenotes}
\end{threeparttable}
\end{table}

\textbf{Results on clean datasets.}
We present the results on clean datasets in \Cref{tab:cifar_clean}. Interestingly, \algo\ consistently outperforms vanilla CE in all scenarios, highlighting the advantages of our proposed soft pseudo-labeling approach. Additionally, our method achieves performance comparable to the widely-used LDAM-DRW baseline. When compared to HAR-DRW, which also addresses class imbalance and label noise, our approach provides an average improvement of over 2\%. Using an auxiliary $1k$ clean and class-balanced validation set, MW-Net achieves performance similar to \algo-DRW in 4 out of 6 cases. However, in the remaining 2 cases, \algo-DRW shows a significant performance gain. Moreover, obtaining such an auxiliary $1k$ clean and class-balanced validation set is challenging in practice, demonstrating the robustness of our method.



\section{Conclusion}
In this paper, we introduce a robust learning framework designed to address both class imbalance and noisy labels. We highlight the limitations of the widely-used small-loss sample selection criterion under long-tailed class distributions and propose a novel small-distance criterion that more accurately identifies correctly-labeled examples across both head and tail classes. Additionally, we introduce a soft pseudo-labeling technique to further enhance training for tail classes. Through extensive experiments on benchmark and real-world datasets, we demonstrate the superiority of our framework compared to existing approaches in both long-tail learning and noisy label detection. We believe that this work will inspire further research into this underexplored yet highly practical task.

While our framework performs well in handling noisy labels, we note that, in the case of clean datasets, the noise detector may still flag certain examples as noisy due to the model's inherent bias toward a two-component Gaussian Mixture Model, leading to the loss of some accurate supervision. Addressing this challenge, by accurately estimating the noise proportion in the training data, presents an interesting direction for future research.

\begin{acknowledgement}
This work was supported by the National Science Foundation of China (62206049, 62225602), the Key Program of Jiangsu Science Foundation (BK20243012), and the Big Data Computing Center of Southeast University. We would like to thank anonymous reviewers for their constructive suggestions.
\end{acknowledgement}


\bibliographystyle{fcs}
\bibliography{reference}

\end{document}